\documentclass[letterpaper, 10 pt, journal, twoside]{IEEEtran}
\usepackage{times}
\usepackage{epsfig}
\usepackage{graphicx}
\usepackage{algorithm}
\usepackage{algpseudocode}
\usepackage{subcaption}
\usepackage{array}
\usepackage{amsmath}
\usepackage{amssymb}
\usepackage{balance}
\usepackage[hidelinks]{hyperref}
\usepackage{booktabs}% http://ctan.org/pkg/booktabs
\usepackage{url}
\newcolumntype{P}[1]{>{\centering\arraybackslash}p{#1}}

\usepackage[flushleft]{threeparttable}
\usepackage{xcolor}
\newcommand{\revised}[1]{\textcolor{black}{#1}}
\begin{document}

\markboth{IEEE Robotics and Automation Letters. Preprint Version. Accepted June, 2022}
{Jiang \MakeLowercase{\textit{et al.}}: A4T}

% \title{\LARGE \bf
% A4T: Hierarchical Affordance Detection for Transparent Objects Depth Reconstruction and Manipulation
% }
\author{Jiaqi Jiang$^{1,2}$, Guanqun Cao$^{1}$, Thanh-Toan Do$^{3}$ and Shan Luo$^{1,2}$ %<-this % stops a space
% \thanks{*This work was not supported by any organization}% <-this % stops a space
\thanks{Manuscript received March 1, 2022; Revised June 4, 2022; Accepted
June 23, 2022. This paper was recommended for publication by Editor Hyungpil Moon upon evaluation of the Associate Editor and Reviewers' comments. This work was partly supported by a University of Liverpool and China Scholarship Council Award, and the EPSRC project ``ViTac: Visual-Tactile Synergy for Handling Flexible Materials" (EP/T033517/1).}
\thanks{$^{1}$smARTLab, Department of Computer Science, University of Liverpool, Liverpool L69 3BX, United Kingdom. Emails: \tt\small\{jiaqi.jiang, g.cao, shan.luo\}@liverpool.ac.uk}
\thanks{$^{2}$J. Jiang and S. Luo are also with Department of Engineering, King's College London, London WC2R 2LS, United Kingdom. E-mails: {\tt\small \{jiaqi.1.jiang, shan.luo\}@kcl.ac.uk}}
\thanks{$^{3}$T.-T. Do is with Department of Data Science and AI, Monash University, Clayton, VIC 3800, Australia. E-mail: {\tt\small toan.do@monash.edu}.}%
\thanks{Digital Object Identifier (DOI): see top of this page.}
}

\title{
A4T: Hierarchical Affordance Detection for Transparent Objects Depth Reconstruction and Manipulation
}

\maketitle
\thispagestyle{empty}
\pagestyle{empty}
%%%%%%%%%%%%%%%%%%%%%%%%%%%%%%%%%%%%%%%%%%%%%%%%%%%%%%%%%%%%%%%%%%%%%%%%%%%%%%%%
\begin{abstract}
% [Background]
Transparent objects are widely used in our daily \revised{lives} and therefore robots need \revised{to} be able to handle them. However, transparent objects suffer from light reflection and refraction, which makes it challenging to obtain the accurate depth maps required to perform handling tasks.
%Transparent objects are widely used in our daily life, e.g., glass dishes and plastic bottles. It would be essential for a robot to detect and grasp transparent objects for performing daily tasks. However, light reflection and refraction, as well as the lack of salient features, make it hard to obtain accurate depth maps of transparent objects to perform such tasks.
In this paper, we propose a novel affordance-based framework for depth reconstruction and manipulation of transparent objects, named \textit{A4T}.
A hierarchical AffordanceNet is first used to detect the transparent objects and their associated affordances that encode the relative positions of an object's different parts. Then, given the predicted affordance map, a multi-step depth reconstruction method is used to progressively reconstruct the depth maps of transparent objects. Finally, the reconstructed depth maps are employed for the affordance-based manipulation of transparent objects.
\revised{To evaluate our proposed method, we construct a real-world dataset \textit{TRANS-AFF} with affordances and depth maps of transparent objects, which is the first of its kind.}
% In the proposed framework, a hierarchical AffordanceNet is first used to detect the objects and their associated affordances. Then given the predicted affordance map, a multi-step depth reconstruction method is used to progressively reconstruct the depth map of transparent objects. Finally, the reconstructed depth map is employed for the affordance-based manipulation of transparent objects.
% To evaluate our proposed method, we collect a real-world affordance dataset of 1,346 RGB-D images of transparent objects.
Extensive experiments show that our proposed methods can predict accurate affordance maps, and significantly improve the depth reconstruction of transparent objects compared to the state-of-the-art method, with the Root Mean Squared Error in meters significantly decreased from 0.097 to 0.042. Furthermore, we demonstrate the effectiveness of our proposed method with a series of robotic manipulation experiments on transparent objects.
\revised{See supplementary video and
results at https://sites.google.com/view/affordance4trans}.
% Detailed improvement:
% [Impact]/[Importance of your work scientifically]
%  
\end{abstract}

%%%%%%%%%%%%%%%%%%%%%%%%%%%%%%%%%%%%%%%%%%%%%%%%%%%%%%%%%%%%%%%%%%%%%%%%%%%%%%%%
\section{Introduction}
\IEEEPARstart{T}{ransparent} objects such as plastic bottles, glass dishes, and windows are widely seen in our daily \revised{lives}. Many of the containers in biomedical and chemical laboratories are also transparent, e.g., Petri dishes, glass flasks, and vials. Humans can expertly identify these transparent objects and interact with them at ease, e.g., picking up a glass cup placed on a table and filling it with water.
% This is a crucial ability for humans in environments with transparent objects.
In recent years, domestic robots have been developed and deployed to assist daily living tasks, and mobile robot chemists have also been created for materials discovery in laboratories~\cite{burger2020mobile}. As transparent objects are widely used in these environments, it is essential that domestic and laboratory robots are able to perceive the transparent objects that are in their surrounding environments and interact with transparent objects safely and dexterously.
\begin{figure}[t]
  \includegraphics[width=\linewidth]{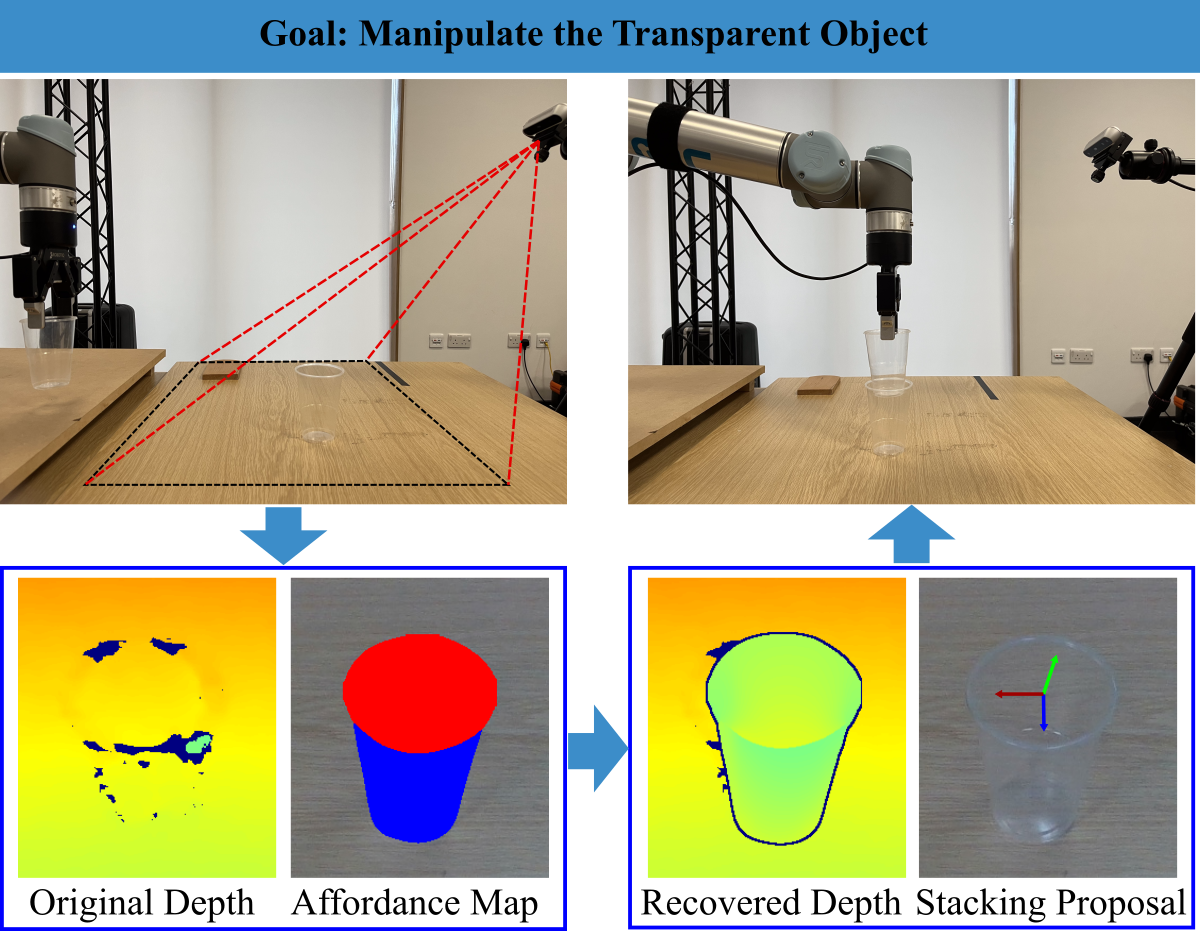}
   \caption{Our goal is to obtain accurate depth maps of transparent objects by leveraging affordance detection, so as to facilitate the manipulation of them, i.e., stacking two plastic cups in the above example. \textbf{Top left:} A plastic cup is in the robot's gripper and another is placed on the table, viewed by an RGB-D camera on its side. \textbf{Bottom left:} A depth map of the cup on the table is obtained from the camera, and its affordance map is predicted, in which the red and blue colours represent the affordance regions with deep cavities to hold liquid (\textit{``contain"}) and that can be held (\textit{``wrap-grasp"}), respectively. \textbf{Bottom right:} The affordance map is used to improve the depth map and predict the center of the gripper for stacking. \textbf{Top right:} With the improved depth maps and the predicted gripping center, the robot can stack the plastic cups successfully.}
\label{fig:overview}
\end{figure}
% \vspace{-0.8cm}

% Challenges 
Many vision-based techniques have been proposed for perceiving poses and shapes of objects in recent decades~\cite{mahler2016dex, murali20206}. However, most of the objects in the previous research have been opaque and the perception of transparent objects remains a challenging problem. Compared to opaque objects, the properties of transparent objects such as reflection, refraction, and the lack of salient features, pose additional challenges to the design of vision algorithms. Their transparent materials violate the Lambertian assumption that optical 3D sensors (e.g., LiDAR and RGB-D cameras) are based on. Hence, most of the depth data from transparent objects are invalid or contain unpredictable noise, which leads to missing or noisy depth approximations of their surfaces. Due to these challenges, most of the current manipulation methods that highly rely on accurate depth information from cameras cannot be applied to the manipulation of transparent objects directly.

%Although many vision-based techniques have been proposed for perceiving the poses and shapes of opaque objects\cite{mahler2016dex}\cite{mahler2017dex}\cite{murali20206}, the perception on transparent objects remains an open and challenging problem. The difficulties mainly lie on the properties of transparent objects . Firstly, they lack salient features, with very little color or texture of their own. Moreover, the properties such as reflection and refraction make transparent objects disobey the geometric light path assumptions used by typical depth sensors like lidar and RGB-D camera, which commonly leads to noisy or missing depth approximations. Hence, most current grasping methods which are highly rely on accurate depth information could not be applied to transparent directly.
%  I have changed this paragraph to only dicussing the drawbacks of ClearGrasp. Then we can give an overview of the difference first in next paragraph. 
 
The existing works divide the manipulation of transparent objects into two steps: depth reconstruction and manipulation planning.
In~\cite{sajjan2020clear}, the depth maps of transparent objects were first reconstructed and then objects were grasped with predicted action possibilities of object regions. However, the depth reconstruction in~\cite{sajjan2020clear} requires reliable initial depth maps of contact edges, i.e., points of contact between transparent objects and tables. This requirement cannot be met for some areas such as regions with deep cavities to contain content and as a result depth maps for such regions cannot be determined.

In this paper, we propose a novel approach named \textit{A4T}, i.e., Affordance for Transparent object depth reconstruction and manipulation. It couples depth reconstruction and manipulation planning for the manipulation of transparent objects via an affordance map that represents the object functionality of each pixel. The affordance map can improve the depth reconstruction and also output the functionalities of object regions for manipulation.
As shown in Fig.~\ref{fig:overview}, a pixel-wise affordance map of object functionalities is first predicted with a hierarchical AffordanceNet that fuses the high-level affordance classification with the low-level fine-grained affordance predictions to avoid the noisy fragments in predicted affordance maps.
%Affordance in this work is considered as at \textit{pixel} level from an image, i.e., a group of pixels which shares the same object functionality is considered as one affordance~\cite{myers2015affordance, nguyen2016detecting,do2018affordancenet}.
% and improve the manipulation of transparent objects in a unified framework, 
% which is the first of its kind. 
With the relative positions of affordance regions encoded, the pixel-wise affordance maps are then used to guide the multi-step depth reconstruction of transparent objects. Furthermore, the reconstructed depth maps and affordance maps are used to improve the affordance-based manipulation.

%In turn

%In this way, the former reconstructed depth can be recognized as a clue for next step reconstruction, so that even the transparent areas enclosed by depth discontinuity boundaries can be reconstructed. 

To evaluate our proposed method, we collected a real-world dataset of transparent objects \textit{TRANS-AFF}, annotated with their affordances and depth maps for opaque objects of the same geometry as the transparent objects to provide the ground truth, which is the very first of its kind. The extensive experimental results demonstrate that our proposed \textit{A4T} can remove the noisy fragments in affordance maps and reconstruct depth maps of transparent objects accurately. Compared to the state-of-the-art method~\cite{sajjan2020clear}, the Root Mean Squared Error in meters is reduced from 0.097 to 0.042. We also conduct the real affordance-based manipulation experiments with a robotic arm. The results show that our proposed depth reconstruction method can significantly enhance the success rates of pouring into transparent objects from 15\% to 80\%.

% Specifically, our contributions of this paper can be summarised as follows:
% \begin{itemize}
% % \item An integrated affordance-based depth reconstruction method has been proposed for manipulation of transparent objects, which is the first of its kind;
% \item We propose a hierarchical AffordanceNet to reduce the noisy fragments via fusing the affordance classification with affordance map prediction;
% \item We propose a multi-step depth reconstruction method that embeds the relative positions of different affordance regions and is aimed for affordance-based manipulation;
% % \item Real robotic manipulation of transparent objects is achieved by using affordance detection and reconstructed depth.
% \item We construct a largest affordance dataset for both affordance detection and depth reconstruction of transparent objects.
% \end{itemize}

The rest of paper is structured as follows:. Section~\ref{Section:2} reviews related works and Section~\ref{Section:3} introduces our A4T framework in detail; Section~\ref{Section:4} details the TRANS-AFF dataset; Section~\ref{Section:5} analyses the experimental results; Finally, Section~\ref{Section:6} summarises the paper and discusses the work.

\section{Related Works}
\label{Section:2}

\subsection{Depth Reconstruction for Transparent Objects Grasping}
Depth reconstruction is an important step towards reliable manipulation of transparent objects as it mitigates the sensor failures in depth images caused by reflection and refraction of transparent objects. Most of the current depth reconstruction methods for transparent objects either rely on a specific capture process or prior knowledge. Han et al.~\cite{han2015fixed} developed a special fixed view capture system, in which parts of the object were immersed in a liquid to make light only reflects for once in objects. In~\cite{qian20163d} and~\cite{phillips2016seeing}, the shapes of transparent objects were recovered with known backgrounds and 3D models, respectively.

There are some other approaches reconstructing object shapes without the prior knowledge about geometries or 3D model. In~\cite{klank2011transparent}, an approach was proposed that first matches pixels from Time-of-Flight images and then reconstructs surfaces with triangulating methods.
Neural radiance fields~\cite{sitzmann2020implicit} were used in~\cite{ichnowski2022dex} to infer the geometries of transparent objects.
% However, it required a long training time for each scene and a set of images with known extrinsics.
Sajjan et al.~\cite{sajjan2020clear} used the global optimisation algorithm proposed in~\cite{zhang2018deep} to reconstruct the missing or noisy depth regions of transparent objects. In~\cite{zhu2021rgb}, Zhu et al. used a local implicit neural representation and an iterative depth refinement model to complete the depth information of transparent objects.
% Instead of only reconstructing the depth information, \cite{xu2022seeing} proposed a joint point cloud and depth completion method to leverage RGB and depth signals of transparent objects.
%  DepthGrasp (IROS 2021)
However, all the above methods only focus on depth reconstruction, and have to be combined with other manipulation planning methods~\cite{mahler2016dex, zeng2018robotic} to manipulate transparent objects. 
% decouple the manipulation of transparent objects into two independent steps: depth reconstruction and manipulation planning.
In our work, we propose a novel approach that couples depth reconstruction and manipulation planning for manipulation of transparent objects via affordance maps that can improve the depth reconstruction and also output the functionalities of object regions for manipulation.

\subsection{Affordance Detection}
The study of affordance learning has attracted increasing attention from researchers in recent years. Currently, there are two main research directions for understanding the affordances of objects, i.e., behavior-grounded affordance learning and pixel-level affordance learning.
The behavior-grounded affordance learning~\cite{jain2011learning,mar2015self,mar2017self} learns object affordance via observing the effects of robot's actions performed on the objects. The tool affordance was defined in~\cite{jain2011learning} with three tool functional features that are annotated by hand. In~\cite{mar2015self}, 2D geometrical visual features were used to represent the functional features for tools. Parallel Self-Organising Maps (SOMs) were used in~\cite{mar2017self} to learn the tool affordance based on their 3D geometry.
% In a recent work~\cite{fang2020learning}, a multi-dimensional continuous action space was created for completing more challenging tasks, instead of a small set of discrete actions.
% drawbacks

The pixel-level affordance learning takes a group of pixels that share the same object functionality as an affordance. In~\cite{myers2015affordance}, the affordances of tool parts were detected from local shapes and geometries. However, those hand-designed features can only capture the information of object properties. In recent years, deep learning approaches have been used to learn affordances and understand the relationship between different parts of one object~\cite{zeng2018robotic,do2018affordancenet,nguyen2016detecting,chu2019recognizing}. An encoder-decoder architecture and Fully Convolutional Networks (FCN)~\cite{long2015fully} were used to efficiently obtain dense affordances predictions in~\cite{nguyen2016detecting} and~\cite{zeng2018robotic}, respectively. Different from previous semantic segmentation methods, in~\cite{do2018affordancenet,chu2019recognizing}, objects were detected (i.e., the object location and the object label) simultaneously with their associated affordances. 
% \cite{chu2019recognizing} treated the affordances as attributes to enhance the learning of visual representations and discarded the attributes during inference.
Instead of only using affordances to enhance the learning of visual representations~\cite{chu2019recognizing}, our hierarchical AffordanceNet fuses the high-level predicted affordance classes and the low-level fine-grained affordance maps to avoid noisy fragments in affordance map.

\begin{figure*}[t]
% 	\fbox{\rule{0pt}{2in} \rule{1\linewidth}{0pt}}
	\includegraphics[width=\textwidth]{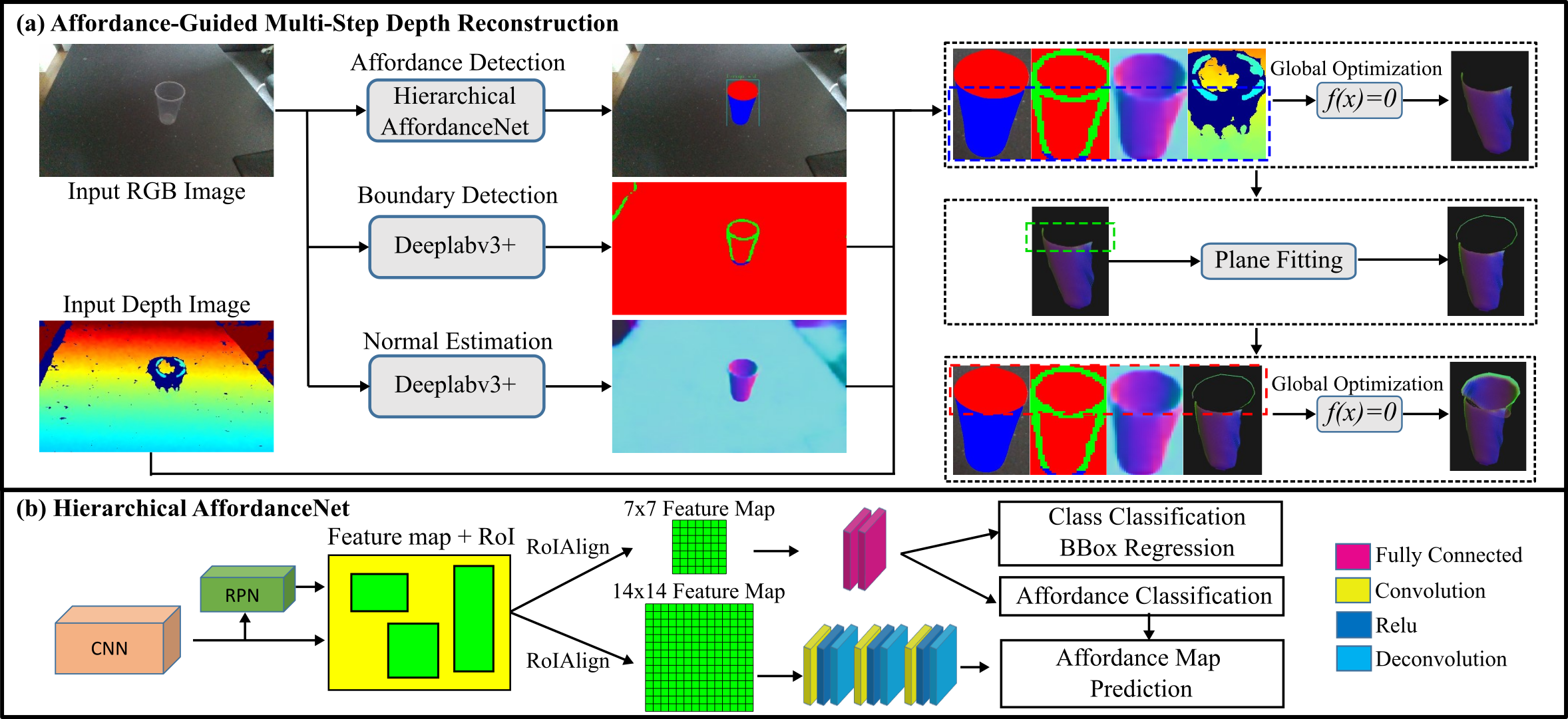}
	\caption{\textbf{(a)}: \textbf{\textit{From Left to Right.}} Given an RGB-D image of a scene with transparent objects, \textbf{A4T} uses three networks to infer 1) affordance maps of transparent objects, 2) occlusion boundaries and contact edges, and 3) surface normals. Then based on the affordance map, the transparent object is progressively reconstructed with a global optimisation method and a plane fitting method. The colourful dash rectangles in the top right corner represent the input information for each reconstruction step.  
	\textbf{(b)}: \textbf{\textit{From Left to Right.}} A deep Convolutional Neural Network (CNN) backbone is used to extract the features of RGB images. The Region Proposal Network (RPN) shares weights with the CNN backbone and outputs RoIs. For each RoI, two RoIAlign layers extract and pool its features to feature maps of a fixed size. Two fully connected layers are used for object classification, object location regression and affordance classification. Three convolutional-deconvolutional layers are used to obtain affordance maps that are fused with the affordance classification scores. Finally, a softmax layer is applied to output a multi-class affordance mask.}
	\label{fig:AffordanceNet}
\end{figure*}

% restore the depth of all areas at once, which results in inaccurate depth maps in the regions without reliable information (for example, the inner surface of a glass cup). In our work, we propose a multi-step depth reconstruction framework to propagate the reliable depth predictions, e.g., the ones on the edges of the cup, which can achieve much more accurate depth predictions for challenging transparent surfaces.

\section{Methodology} 
\label{Section:3}
In this work, we propose an affordance-based depth reconstruction framework that facilitates the robotic manipulation of transparent objects, as shown in Fig.~\ref{fig:AffordanceNet}. 
The transparent object in the image is first detected, with its affordance classes and affordance map, using a novel hierarchical AffordanceNet. Then an affordance-guided multi-step depth reconstruction method is used to progressively reconstruct the depth maps of the transparent object. Finally, using the reconstructed depth the robot executes the affordance-based manipulation tasks, i.e., picking, pouring, and stacking.

% \begin{figure*}[t]
% 	\centering
% % 	\fbox{\rule{0pt}{2in} \rule{1\linewidth}{0pt}}
% 	\includegraphics[width=1.9\columnwidth]{Figures/Fig3.png}
% 	\caption{An overview of our multi-step depth reconstruction method.}
% 	\label{fig:multi-step}
% \end{figure*}

\subsection{Hierarchical AffordanceNet}\label{section:IIIA}
Our hierarchical AffordanceNet extends our previous AffordanceNet~\cite{do2018affordancenet} by \revised{introducing an affordance classification module, as shown in Fig.~\ref{fig:AffordanceNet}-(b). By fusing the high-level affordance classes with the low-level fine-grained affordance map, our hierarchical AffordanceNet can make the predicted affordance map consistent with the predicted affordance classes.}
The RGB features are first extracted, with the ResNet50~\cite{he2016deep} network as the backbone. \revised{A Region Proposal Network (RPN)} that shares the same weights with the convolutional backbone is then used to generate the \revised{Regions of Interest (RoIs)}. They are followed by two kinds of RoIAlign~\cite{he2017mask} layers that extract and pool the corresponding features of candidate Bounding Boxes (BBoxes) into 7$\times$7 feature maps (for the BBox regression and affordance classification) and 14$\times$14 feature maps (for the affordance map prediction).
% , as the bigger feature map size can enhance the affordance predictor performance.

\textbf{Affordance classification.}
The affordance classification is recognised as a multi-label classification problem, and parallel to the object detection. We feed the 7$\times$7 feature maps to two fully connected layers, each with 1024 neurons, followed by $N$ binary classifiers, where $N$ is the number of affordance classes. The $N$ binary classifiers output a N-dimensional classification score $A_c$ that represents the possibility  of each affordance within the proposal.
To train the affordance classification, a multinomial cross entropy loss $L_{AFF-C}$ is used and defined as follows:
\begin{equation} \label{equ1}
\begin{aligned}
L_{AFF-C} = -\frac{1}{N}\sum_{i=0}^{N}(y_i*\revised{\log}(A_c(i))\\
+(1-y_i)*\revised{\log}(1-A_c(i)))
\end{aligned}
\end{equation}
where $y_i$ denotes the label of the $i^{th}$ affordance class.

\textbf{Affordance map prediction.} 
As for the pixel-wise affordance map prediction, the 14$\times$14 feature maps are upsampled to 112$\times$112 with three convolutional-deconvolutional layers. Then a convolutional layer is used to generate the output affordance map $A_m$ with $N+1$ channels, where the plus one represents the additional channel for background.
To train the affordance map, a multinomial cross entropy loss $L_{AFF-M}$ is used and computed as follows:
\begin{equation} \label{equ2}
\begin{aligned}
L_{AFF-M} = -\frac{1}{K}\sum_{i\in RoI}\revised{\log}(m^i_{s_i})
\end{aligned}
\end{equation}
where $m^i_{s_i}$ is the softmax output at pixel $i$ for the true label
${s_i}$, and $K$ is the number of pixels in the RoI.

% All the convolutional layers in the convolutional-deconvolutional layers have zero padding $d = 1$, stride $s = 1$, and a kernel size $S_f = 3$ to keep the size of the feature map. The filters in all the deconvolutional layers have a size of 2$\times$2, with zero padding $d = 0$ and stride $s = 2$, which can double the size of the feature map based on the following equation:
% \begin{equation} \label{equ2}
% S_{o}=s *\left(S_{i}-1\right)+S_{f}-2 * d
% \end{equation}
% where $S_i$ and $S_o$ are the sizes of the input feature map and the output feature map, respectively; $S_{f}$ is the filter size; $s$ and $d$ are stride and padding parameters, respectively.

\textbf{Fusion of affordance classification.} Different from \cite{chu2019recognizing} that only uses affordance classification to guide feature learning during training, we fuse the affordance classification scores $A_c$ and the predicted affordance map $A_m$ to infer the final pixel-wise affordance map $A_m^F$:
\begin{equation} \label{equ3}
A_m^F = [1, A_c]\otimes A_m
\end{equation}
where $A_m^F$ with $N+1$ channels and $\otimes$ is the element-wise product, \revised{also known as Hadamard product}. Note that the value 1 in Equation \ref{equ3} is used to keep the background channel of the pixel-wise affordance map. After obtaining the final affordance map $A_m^F$, a high-resolution affordance mask is then generated by a softmax layer.

\subsection{Affordance-guided Depth Reconstruction}\label{Sec:III:B}
In this subsection, we propose a multi-step depth reconstruction method that embeds the relative positions of different affordance regions. 
\revised{Different from~\cite{sajjan2020clear} that reconstructs the entire image at one step, we reconstruct every individual object with a multi-step method that reconstructs different affordance regions one by one. In this way, the reconstructed depth of the current affordance region can be recognised as a clue for the reconstruction of the next affordance region, so that even the areas enclosed by depth discontinued boundaries can be reconstructed. Firstly, the RGB-D image, the predictions of surface normals and occlusion boundaries are cropped for each detected instance. Then the depth map of one transparent object is progressively reconstructed with a global optimisation method and a RANSAC-based plane fitting method. 
}
%  To recover the depth information of transparent objects more accurately, a multi-step depth reconstruction method is proposed in this subsection. Our method only focus on each transparent objects separately, and reconstruct its different parts individually based on the predicted affordance map from Sec. \ref{section:IIIA}. The overview of this approach is shown in Fig. \ref{fig:multi-step} and the details will be provided in the following paragraphs.

% Firstly, the transparent objects, as well as their corresponding surface normals and contact edges predicted by the pre-trained DeepLabv3+ networks in \cite{sajjan2020clear} are extracted given the predicted attributes and locations from Sec. \ref{section:IIIA}. Then the pixel-wise affordance map is used to remove the remove unreliable depth measurements from the depth camera and revised the contact edges so that . Finally the depth pipeline proposed by Zhang and Funkhouser\cite{zhang2018deep} and a RANSAC-based edge completion method are utilized to reconstruct the depth information in the surface area and edge respectively. The first optimisation algorithm fills in the removed depth using the predicted normals to guide the shape of the reconstruction, while observing the depth discontinuities indicated by the occlusion boundaries.

% which is used to find the contact edges in affordance occlusion boundary
\textbf{Cropping of Transparent Objects.} Following~\cite{sajjan2020clear}, first two DeepLabv3+ networks~\cite{chen2018encoder} are used to predict surface normals and boundaries of transparent objects. The boundaries include both occlusion boundaries and contact edges \revised{where depths are not continued and objects are contacted with tables, respectively. 
To reconstruct every individual object instead of the entire image in~\cite{sajjan2020clear}, the depth image, as well as its corresponding surface normals image and occlusion boundary image are cropped.}
Then based on the predicted affordance map from Sec.~\ref{section:IIIA}, the unreliable depth measurements of transparent objects are removed from the depth image.

% and the predicted occlusion boundaries from DeepLabv3+ are replaced with the contour of each affordance part.
% % to make the occlusion boundary consistent with the depth-removed regions.
% In this way, \shan{the predicted contact edges from DeepLabv3+ may not lie on the replaced occlusion boundary obtained from affordance detection. Hence, a distance thresholding method is used to determine whether the pixels on the occlusion boundary belong to the contact edge. If any pixel is less than a predefined threshold from the contact edge, it will be treated as a part of the contact edge.}

\textbf{Depth reconstruction.}
For each detected transparent object, we fill in their removed depth map with two methods, i.e., the global optimisation algorithm proposed by Zhang and Funkhouser~\cite{zhang2018deep} and a RANSAC-based plane fitting method~\revised{\cite{yang2010plane}}.
Specifically, the global optimisation algorithm completes the depth information of the transparent objects by minimizing the integrated error $E$:
\begin{equation}
E=\lambda_{D} E_{D}+\lambda_{S} E_{S}+\lambda_{N} E_{N} B
\end{equation}
where $E_{D}$ measures the distance between the estimated depth and the observed raw depth, $E_{S}$ measures the difference between the depths of neighboring pixels, and $E_{N}$ measures the consistency between the estimated depth and the predicted surface normal, respectively; $\lambda_{D}$, $\lambda_{S}$, $\lambda_{N}$ are chosen according to~\cite{sajjan2020clear,zhang2018deep,liu2020depth}. \revised{$\lambda_{D}$ is set to 1000 to encourage the reconstructed depth as close as possible to the reliable raw depth of opaque objects; $\lambda_{S}$ is set to 0.001 to encourage adjacent pixels to have the same depths; $\lambda_{N}$ is set to 1 as a reference.} $B$ down-weights the normal terms based on the predicted probability that a pixel is on an occlusion boundary.

The RANSAC-based plane fitting method is used to complete the edge depth with part of known points on it. Firstly, it randomly selects \revised{three} points and uses Singular \revised{Value} Decomposition (SVD) to initialize the plane parameters~\revised{\cite{vidal2008multiframe}}. Then it detects all points belonging to the calculated plane, according to a given threshold. Afterwards, it repeats these procedures $I$ ($=500$ in our case) times. In each time, the plane parameters will be updated if the current result is better than the last saved one. If the objects vertically are placed on a flat table, we constrain the fitted plane parallel to the table to obtain a more robust fitting result.

Firstly, we use the global optimisation method to complete the affordance regions with contact edges on their boundaries. Then the following reconstruction processes will be determined by the connected edges on first reconstructed regions (i.e., points of contact between two regions of different affordances). If the connected edge is depth discontinued, such as the connected edge between ``wrap-grasp" and ``contain" \revised{(the blue region and the red region on cropped affordance map in Fig.~\ref{fig:AffordanceNet}, respectively)}, the RANSAC-based plane fitting method would be used to complete the boundary depth of their neighbor affordance regions. After that, the global optimisation method is used to complete their neighbor affordance regions, with the optimised depth on boundaries taken as the reliable depth. Otherwise, if the connected edge is depth continued, such as the connected edge between ``wrap-grasp" and ``support" (the bottom of containers that can be supported by a table), the global optimisation would be used again to reconstruct the depth in their neighbor affordance regions.

% \begin{algorithm}[H]
%     \caption{Multi-Step Depth Reconstruction}
%     \begin{algorithmic}
%         \Require 
%           $I_{rgb}$:  Original RGB image
%           $I_d$: Original depth image
%           $A$:  Detected Affordance Map
%         \Ensure
%           $O$: Reconstructed depth image
%         \State  surface$\_$normals, contact$\_$edges $\gets$ DeepLabv3+($I_{rgb}$)
%         \State  affordance$\_$prioritization(contact$\_$edges)
%         \State  
%         \If {$k=0$}
%           \State $k\gets 1;$  $n_1 \gets 1;$
%           \State $c_1 \gets p_i;$ $y_1 \gets yaw_i;$
%       \Else

%       \EndIf
%         \label{code:recentEnd}
%     \end{algorithmic}
    
% \end{algorithm}

\subsection{Affordance-based Manipulation}
\begin{figure}[H]
	\centering
% 	\fbox{\rule{0pt}{2in} \rule{1\linewidth}{0pt}}
	\includegraphics[width=1\columnwidth]{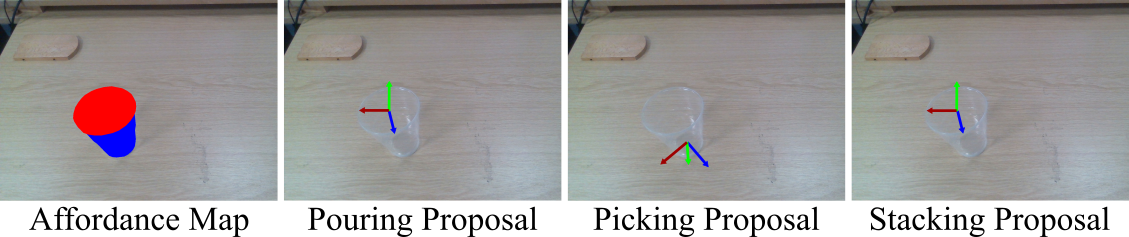}
	\caption{An example of the generated proposals for different tasks. Red, green, and blue arrows represent the x-axis, y-axis, and z-axis of the end-effector.}
	\label{fig:sketch}
\end{figure}

To demonstrate that our method has the ability to manipulate the transparent objects, we take three manipulation tasks, picking, pouring and stacking as examples, as shown in Fig.~\ref{fig:sketch}.

% (Our method will be similar to cleargrasp method)

\textbf{Pouring.} In the pouring task, the ``contain" affordance is used to generate the pouring proposal $P_{pour} = [R, T]\in SE(3)$. Specifically, the translation vector $T$ is calculated by averaging the point clouds on ``contain" region's boundary. As for the rotation vector $R$, the z-axis of the end-effector is perpendicular to the plane where the ``contain" boundary lies on and is in the same vertical plane with y-axis. If the robotic arm pours with a container, an offset of container's length will be added to the y-axis.
% (Our method will be much better than cleargrasp method)

\textbf{Picking.} In the picking task, the ``wrap-grasp" affordance is used to generate the picking proposal $P_{pick} = [R, T]\in SE(3)$. Specifically, the translation vector $T$ is calculated by averaging the point clouds of ``wrap-grasp" region. As for the rotation vector $R$, the z-axis of the end-effector is parallel to the predicted surface normal of the center pixel of ``wrap-grasp" region and is in the same vertical plane with y-axis.

\textbf{Stacking.} In the stacking task, the ``contain" and ``support" affordances are used to generate the stacking proposal $P_s = [R, T]\in SE(3)$, where the stacking proposals for objects containing ``support" affordance and ``contain" affordance are generated with the same ways to generating the picking proposal and pouring proposal, respectively. 

\section{Dataset Collection}
\label{Section:4}
\begin{figure}[t]
   \includegraphics[width=1\linewidth]{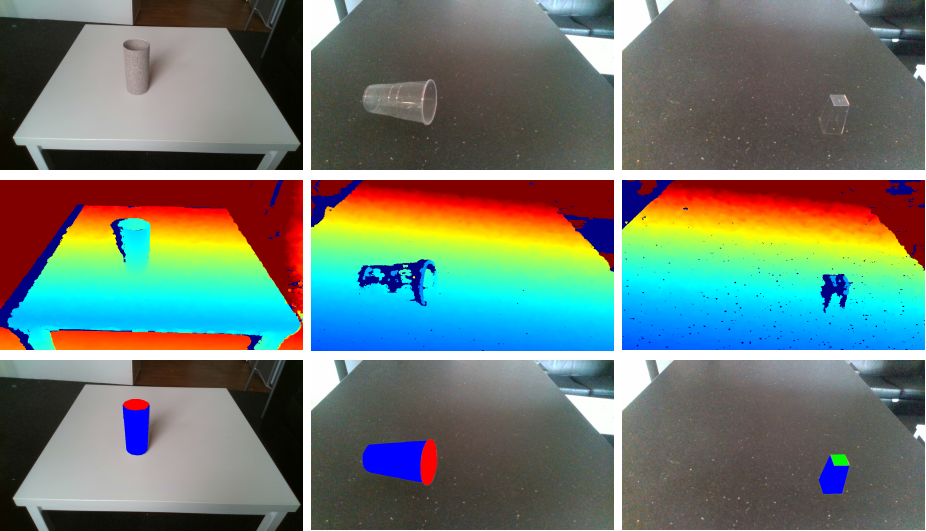}
   \caption{Sample objects from the \textbf{TRANS-AFF} Dataset. \textbf{Top row}: RGB images. \textbf{Middle row}: Depth images.  \textbf{Bottom row}: Pixel-wise affordances label. Red, blue and green colour represent ``\textit{contain}'', ``\textit{wrap-grasp}'' and ``\textit{support}'' respectively. }
\label{fig:samples}
\end{figure}
\subsection{TRANS-AFF Dataset}\label{TRANS-AFF}
\begin{figure}[t]
% \fbox{\rule{0pt}{1.4in} \rule{0.9\linewidth}{0pt}}
  \includegraphics[width=1\linewidth]{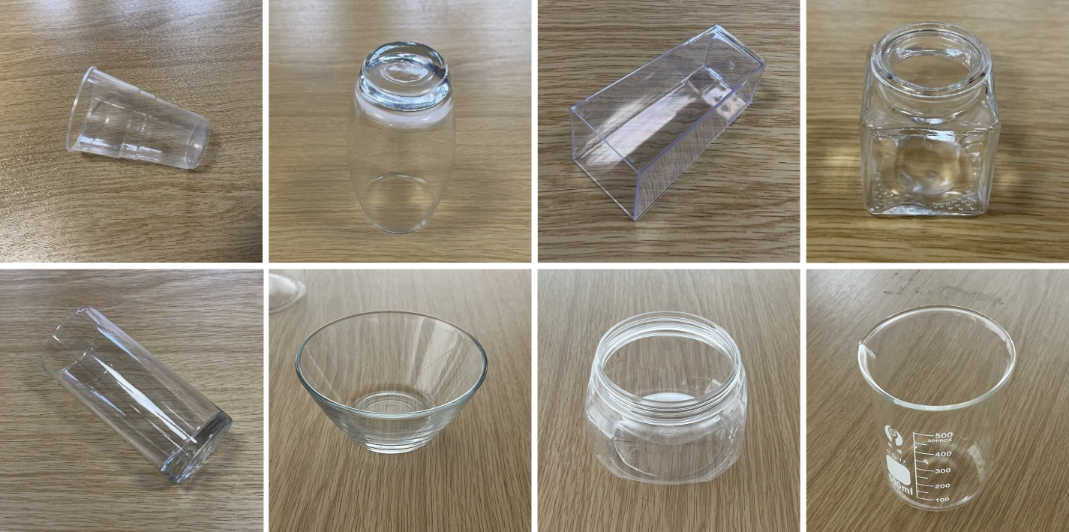}
   \caption{Objects in the \textbf{TRANS-AFF} Dataset. \textbf{First row} (from left to right): disposable cup, highball cup, rectangular cup, and square jar; \textbf{Second row} (from left to right): tumble cup, bowl, round jar, and measurement cup.}
\label{fig:objects}
\end{figure}

To evaluate our proposed methods, we collect a new RGB-D Affordance dataset for transparent objects named as \textbf{TRANS-AFF}, with some samples shown in Fig.~\ref{fig:samples}. We collect the dataset with a RealSense D435i camera and a RealSense D415 camera that have a resolution of 1280$\times$720 pixels and a resolution of 1920$\times$1080 pixels, respectively. To provide accurate ground-truth depth maps for the transparent objects, we collected a ``twin" RGB-D pair for every scene that contains transparent objects. In the ``twin" setup, the transparent objects in the original image are replaced with an identical spray-painted instance that can reflect light evenly and provide accurate depth information. 
In total, there are 1,346 pairs of RGB and depth images for 8 graspable containers \revised{as shown in Fig.~\ref{fig:objects}}. Compared to other affordance datasets in the literature~\cite{myers2015affordance,nguyen2016detecting}, our \textbf{TRANS-AFF} dataset is the first one that includes transparent objects and provides accurate depth maps. Furthermore, to our best knowledge, it is also the first dataset designed for both depth reconstruction and affordance detection.

In this dataset, we define three surface’s effective affordances by the way it comes in contact with the objects it affects. For example, a water cup standing upright on the table has two affordance parts, i.e., the inner surface and the outer surface. The inner surface of a cup has the effective affordance ``\textit{contain}'', as it comes in contact with the liquid that is contained. The outer surface of a cup has the affordance ``\textit{wrap-grasp}'' as it can be held with the hand and the palm. In addition to the two above, we set the bottom of the cup as ``\textit{support}'', as it can make the cup stand upright on the table. 
Those three affordances are summarised in Table~\ref{tab:aff}.

\revised{We split all the data in the dataset instead of the objects by a ratio of 7:2:1 for training, validation and testing. In this way, all the transparent objects used in this work are known a priori, which is consistent with the fact that the models of the objects used in laboratories and factories are available from their manufacturers.} \revised{To reduce the influence of different shadows and caustics between opaque and transparent twins, only transparent objects are used for training the hierarchical AffordanceNet to focus on the features of transparent objects.}  

\begin{table}
\centering
    \caption{Description of the three affordance labels (see Fig.~\ref{fig:samples} for examples).}
    \label{tab:aff}
    \scalebox{0.9}{
    \begin{tabular}{c||c}
        \hline \text { Affordance } & \text { Description } \\
        \hline \text { contain } & \text { With deep cavities to hold liquid or other content. } \\
        \hline \text { wrap-grasp } & \text { Regions that can be held with the hand. } \\
        \hline \text { support } & \text {A flat part that can be supported vertically on a table. } \\
    \hline
    \end{tabular}}
\end{table}

\begin{figure*}
  \includegraphics[width=1\linewidth]{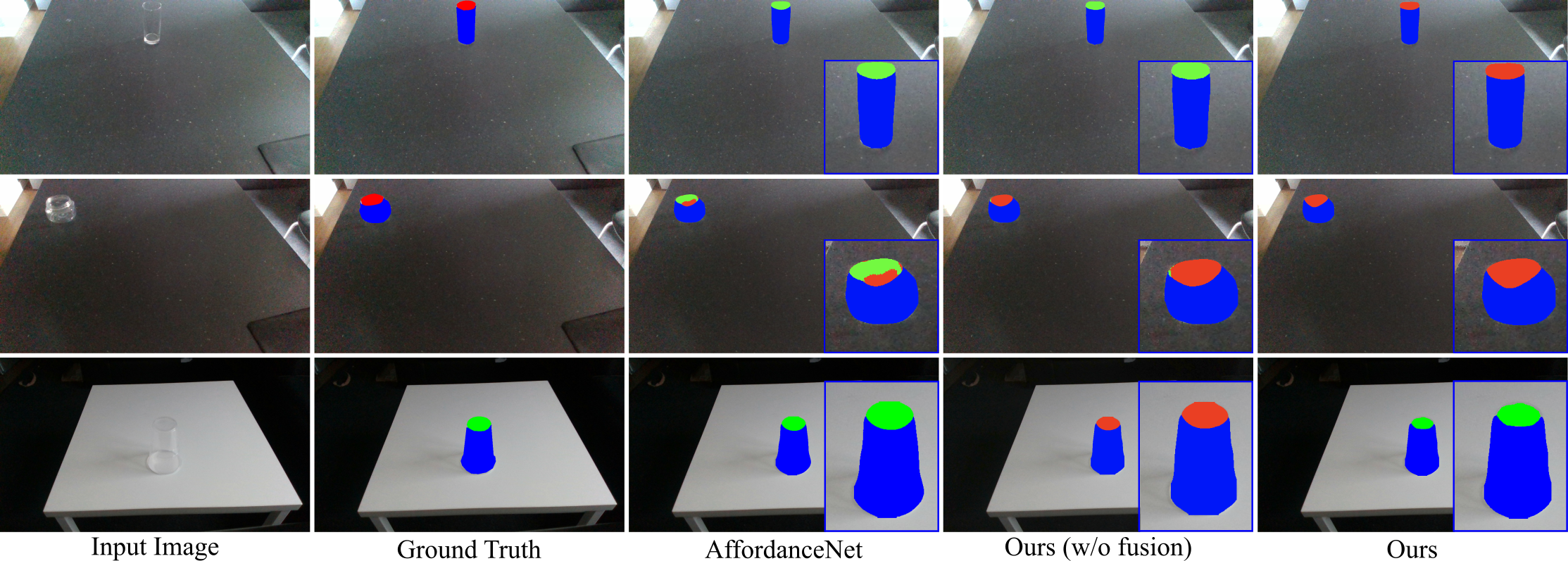}
   \caption{Comparisons of affordance detection. The first two rows show that our method outperform the original AffordanceNet~\cite{do2018affordancenet} and can avoid the noisy fragments in affordance maps. The third row shows that our hierarchical AffordanceNet may lose some fine-grained information as it involves the high-level affordance classification.}
\label{fig:baseline}
\end{figure*}

\begin{table}
	\centering
		\caption{The ablation study of the hierarchical AffordanceNet}
		\label{tab:ablationstudy}
        \scalebox{1}{
		\begin{tabular}{c c | c |c |c | c }
			\hline
			 AFF-C & AFF-F  & contain & support & grasp & Average \\
			\hline
			\hline
% 			{=|=|=|=|=|=|=}
             & &    83.80 & \textbf{85.08}& \textbf{95.26} & 88.04 \\
            \checkmark & & 83.95 & 83.39 & 94.75& 87.36 \\
    		\checkmark &\checkmark & \textbf{87.94} & 83.48 & 94.92 & \textbf{88.78}\\
    		
    		\hline
		\end{tabular}}
\end{table}

\section{Experiment Results}
\label{Section:5}
% To prove our framework, 
% \shan{Here goes an overview of the experiments}
In this section, we conduct a series of experiments to evaluate our affordance detection for transparent object depth reconstruction and manipulation. The goal of the experiments are three-fold: 1) To evaluate the effectiveness of the proposed hierarchical AffordanceNet in our \textbf{TRANS-AFF} dataset; 
% \guan{To evaluate the effectiveness of affordance detection with our proposed hierarchical AffordanceNet using real-world datasets; }
2) To investigate how affordance detection can improve the performance of transparent object's depth reconstruction; 3) To investigate how the reconstructed depth information can improve the success rate of affordance-based manipulation.

\subsection{Multi-Affordance Prediction}
To evaluate the affordance detection results, the weighted F-measures metric $F_{\beta}^{\omega}$ is used and can be computed as follows:
\begin{equation}
F_{\beta}^{\omega}=\left(1+\beta^{2}\right) \frac{\operatorname{Pr}^{\omega} \cdot R c^{\omega}}{\beta^{2} \cdot \operatorname{Pr}^{\omega}+R c^{\omega}}
\end{equation}
where $\beta = 1$, $\operatorname{Pr}^{\omega}$ and $\operatorname{Rc}^{\omega}$ are the weighted precision and recall values, respectively. All the networks are trained for 20 epochs using a fixed learning rate of 0.001 with the SGD optimizer. 

Table~\ref{tab:ablationstudy} summarizes the average $F_{\beta}^{\omega}$ score of the aforementioned networks when testing with the images that have a resolution of 1280$\times$720 pixels in TRANS-AFF dataset. The ``AFF-C" and ``AFF-F" in~Tab. \ref{tab:ablationstudy} represent the affordance classification module and the fusion of affordance classification, respectively. The results show that our hierarchical model can significantly improve the performance on affordance ``contain," and has comparable performance to AffordanceNet.

We also show the qualitative results of affordance detection in Fig.~\ref{fig:baseline}. The ``AFF-C" can enhance the learning of affordance representation but cannot fully avoid noisy fragments, as the second row of Fig.~\ref{fig:baseline}. Via fusing the high-level affordance classes with the low-level fine-grained affordance map, our hierarchical AffordanceNet can avoid the wrong affordance class and the noisy fragments as the first row and the second row of Fig.~\ref{fig:baseline}, respectively. However, there is a trade-off between affordance classification and pixel-wise affordance segmentation. The affordance classification module sometimes influences the \revised{fine-grained} affordance segmentation, as the third row of Fig. \ref{fig:baseline}, which limits the overall performance.

% \begin{table}[h]
% 	\centering
% 		\caption{The effect of the size of the affordance map on the detection accuracy}
% 		\label{tab:size}
%         \scalebox{1}{
% 		\begin{tabular}{c| c | c | c |c}
% 			\hline
% 			& contain & support & grasp & Average \\
% 			\hline
% 			\hline
% % 			{=|=|=|=|=|=|=}
%     		28$\times$28     &  &  & \\
%     		56$\times$56  &  &  & \\
%     		112$\times$112    & & & \\
%     		\hline
% 		\end{tabular}}
% \end{table}

%train 和 test的区别在于，

% designed a simple labeling interface that prompts users to manually annotate suction and grasp proposals over RGB-D images collected from the real system. For

\begin{figure*}[t]
  \includegraphics[width=1\linewidth]{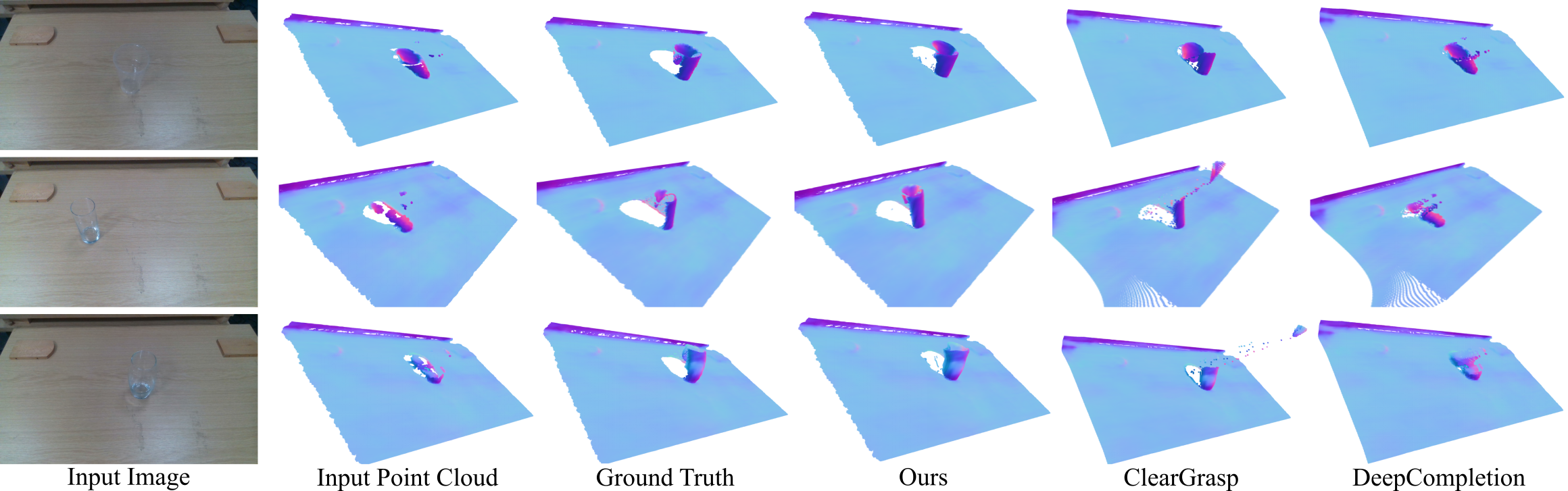}
   \caption{Samples of baseline comparison. We view those point clouds from different orientation to clearly compare the reconstruction of the ``contain" regions. As shown in this figure, our method can achieve stable reconstruction of transparent objects, especially the inner surfaces of cups.}
\label{fig:depth_baseline}
\end{figure*}

% Next,  we  would  like  to  produce  an  identical  architecturethat uses depth input in the form of an HHA encoding [20](which  encodes  a  depth  image  geocentrically  using  threechannels:  horizontal  disparity,  height  above  ground,  andangle  between  the  pixel’s  local  surface  normal  and  theinferred  gravity  direction).

\subsection{Depth Reconstruction}
For the depth estimation, we use the standard metrics used in previous studies~\cite{eigen2014depth}: Root Mean Squared Error in meters (RMSE), the median error relative to the depth (Rel), Mean Absolute Error (MAE) and percentage of pixels with predicted depths falling within an interval ($[\delta=\max ($predicted$/$true$,$true$/$predicted$)]$, where $\delta$ is 1.05, 1.10 or 1.25)~\cite{sajjan2020clear}.

We compare our proposed algorithm with the state-of-the-art algorithm ClearGrasp~\cite{sajjan2020clear} and DeepCompletion~\cite{zhang2018deep} on the images within our TRANS-AFF dataset.
The depth reconstruction results on different affordance regions are shown in Table~\ref{tab:re3}. It is noticed that our method has a similar performance on ``wrap-grasp" regions and ``support" regions to ClearGrasp method. However, there is a significant improvement on ``contain" regions with the RMSE in meter decreased from 0.182 to 0.046, which contributes to the notable RMSE decrease for all regions from 0.097 to 0.046. 

The reason why ClearGrasp method performs worst in ``contain" regions is because the global optimisation method cannot determine the depth of the regions that are enclosed by occlusion boundaries.
% Similarly as shown in the fifth column of Fig. \ref{fig:depth_baseline}, the inner portions of the containers have wrong depth information that is reconstructed with ClearGrasp method.
However, our algorithm can achieve a stable depth reconstruction on those challenging areas as shown in the fourth column of Fig. \ref{fig:depth_baseline}. 
It is also noticed that our method shows the worst accuracy on ``support" regions compared to ``contain" and ``wrap-grasp," this is because the DeepLabv3+ network incorrectly predicts the surface normals of some objects in the ``support" regions as the surface normals of the ``contain" regions. 
% \emph{dd}

\begin{figure}[t]
   \includegraphics[width=1\linewidth]{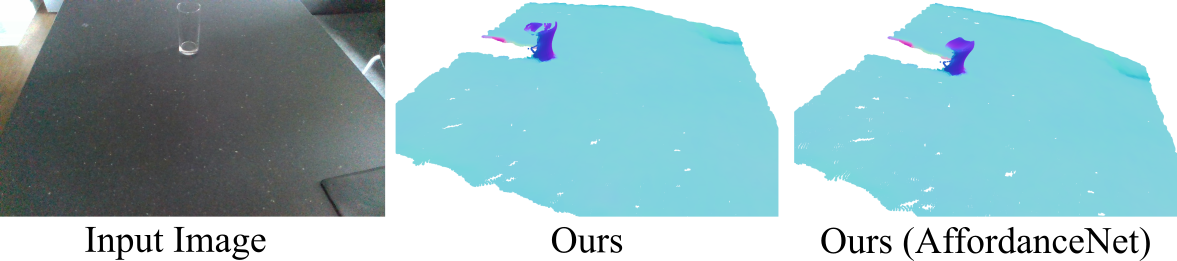}
   \caption{Comparison using different AffordanceNet. Using the affordance prediction from AffordanceNet~\cite{do2018affordancenet} results the failed reconstruction of the ``contain" region.    }
\label{fig:netcomparison}
\end{figure}

As shown in Fig.~\ref{fig:netcomparison}, we also compare the reconstructed point clouds of the first picture of Fig.~\ref{fig:baseline} when different affordance networks are used. Our method shows slightly better performance compared to using the original AffordanceNet~\cite{do2018affordancenet}, as the mispredicted affordance map leads to the wrong reconstruction of the ``contain" region. More importantly, our hierarchical AffordanceNet can predict the affordance map without noisy fragments as the second row of Fig.~\ref{fig:baseline}, which makes the planning of our multi-step depth reconstruction easier.
\revised{Objects other than graspable containers do not have ``contain" affordance regions as they do not have cavities to contain content. This makes the plane fitting method in our multi-step reconstruction unneeded and hence our multi-step reconstruction will degenerate into ClearGrasp~\cite{sajjan2020clear}. The objects such as bath bomb molds without ``contain" affordance regions were tested in ClearGrasp~\cite{sajjan2020clear} and could be reconstructed well with the vanilla global optimisation method in~\cite{zhang2018deep}.}
% \revised{It should be noted that our method can work well for the objects beyond cups such as spherical objects. This is because for the objects without areas enclosed by depth discontinuity boundaries, i.e. the ``contain" region, our multi-step reconstruction shows a similar performance to  ClearGrasp~\cite{sajjan2020clear} that works well on the object with different geometries.
% Cups were chosen for our experiments, as they can significantly demonstrate the superiority of our method.}
% 

% \begin{table*}[h]
% 	\centering
% 		\caption{Comparisons on depth reconstruction of contain regions}
% 		\label{tab:re3}
%         \scalebox{0.9}{
% 		\begin{tabular}{c| c | c | c |c | c| c }
% 			\hline
% 			 & RMSE$\downarrow$ & Rel$\downarrow$  & MAE$\downarrow$ & $\delta_{1.05} \uparrow$ &$\delta_{1.10} \uparrow$&$\delta_{1.25} \uparrow$\\
% 			\hline
% 			\hline
% % 			{=|=|=|=|=|=|=}
%             DeepCompletion\cite{zhang2018deep} \\
%     		ClearGrasp\cite{sajjan2020clear}    \\
%     		Ours$_{\rm AFF}$   \\
%     		Ours$_{\rm HAFF}$  \\
%     		\hline
% 		\end{tabular}}
% \end{table*}

\begin{table*}
	\centering
		\caption{Baseline comparisons on different evaluation regions}
		\label{tab:re3}
        \scalebox{1}{
		\begin{tabular}{c| c| c | c | c |c | c| c }
			\hline
			& Evaluation Region & RMSE$\downarrow$ & Rel$\downarrow$  & MAE$\downarrow$ & $\delta_{1.05} \uparrow$ &$\delta_{1.10} \uparrow$&$\delta_{1.25} \uparrow$\\
			\hline
			\hline
% 			{=|=|=|=|=|=|=}
            DeepCompletion\cite{zhang2018deep}& All & 0.105 & 0.162 & 0.080
            &24.81 & 38.35 &79.24\\
    		ClearGrasp\cite{sajjan2020clear}& All & 0.097 & 0.139 & 0.072 & 53.51 &64.44 & 79.46
 \\
    % 		Ours$_{\rm AFF}$   \\
    		Ours&  All &\textbf{0.042}&\textbf{0.064}&\textbf{0.033}&\textbf{63.33}&\textbf{86.47}&\textbf{95.78}\\
    		\hline
            DeepCompletion\cite{zhang2018deep}& contain & 0.106 & 0.185 & 0.096 & 15.40 &24.83 & 70.67 \\
    		ClearGrasp\cite{sajjan2020clear}& contain & 0.182 & 0.329 & 0.172 & 14.20 &19.70 & 41.77 \\
    % 		Ours$_{\rm AFF}$   \\
    		Ours& contain &\textbf{0.046}&\textbf{0.056}&\textbf{0.032}&\textbf{63.50}&\textbf{85.16}&\textbf{93.66} \\
    		\hline
    		DeepCompletion\cite{zhang2018deep}& wrap-grasp & 0.089 &0.141&0.069&28.31 &43.03&84.89 \\
    		ClearGrasp\cite{sajjan2020clear}& wrap-grasp &0.036 &\textbf{0.049} &\textbf{0.025}&\textbf{72.46} &86.28&96.43\\
    % 		Ours$_{\rm AFF}$   \\
    		Ours& wrap-grasp &\textbf{0.034}&0.059& 0.030&66.39&\textbf{88.98}&\textbf{96.96}  \\

    		\hline
    		DeepCompletion\cite{zhang2018deep}& support & 0.139 & 0.192 & 0.090& 35.97& 49.81&70.60\\
    		ClearGrasp\cite{sajjan2020clear}& support &0.066&\textbf{0.090}&\textbf{0.044}&\textbf{60.07}&73.26&87.44 \\
    % 		Ours$_{\rm AFF}$   \\
    		Ours& support &\textbf{0.060}&0.100&0.048&45.86&\textbf{78.19}&\textbf{91.91}  \\
    		\hline
		\end{tabular}}
\end{table*}

\begin{table}
	\centering
		\caption{Comparison on success rate of different manipulation tasks. }
		\label{tab:manipulation}
        \scalebox{1}{
		\begin{tabular}{c| c | c | c }
			\hline
			Depth Reconstruction& picking & pouring & stacking \\
			\hline
			\hline
% 			{=|=|=|=|=|=|=}
    		 \revised{RealSense}   & 20.0\% &  0.0\%& 25.0\%  \\
    		ClearGrasp  & \textbf{82.5}\% & 15.0\% & 57.5\% \\
    		Ours    & \textbf{82.5}\% & \textbf{80.0}\% & \textbf{87.5}\%\\
    		\hline
		\end{tabular}}
\end{table}

\subsection{Real-world Manipulation}
For the manipulation tasks, we use the average success rate = $\frac{\#successful\ attempts}{\#total\  attempts}$ as the evaluation metric.
A picking attempt will be recognised as a successful case if the object is grasped and lifted up for 5 seconds. A pouring attempt will be recognised as a successful case if the object in the robotic hand or the content in the grasped container falls into the container. A stacking attempt will be recognised as a successful case if the object in robotic hand is stacked with the same object on the table. 

We conduct the affordance-based  manipulation experiments with a robotic system that consists of a 6-DoF robotic manipulator UR5 from Universal Robots, a Robotiq-85 parallel two-finger gripper, and an Intel RealSense D415 camera. We use an ArUco Marker to achieve hand-eye calibration. Every manipulation task is tested with 40 attempts, and the results are summarised in Table~\ref{tab:manipulation}. ``RealSense" represents using the original depth information obtained from the RealSense camera to generate manipulation proposals. 
We set all objects vertically on the table and add a protection height to avoid the collision between the robotic arm and the table when manipulating the transparent objects.

\begin{figure}[t]
  \includegraphics[width=1\linewidth]{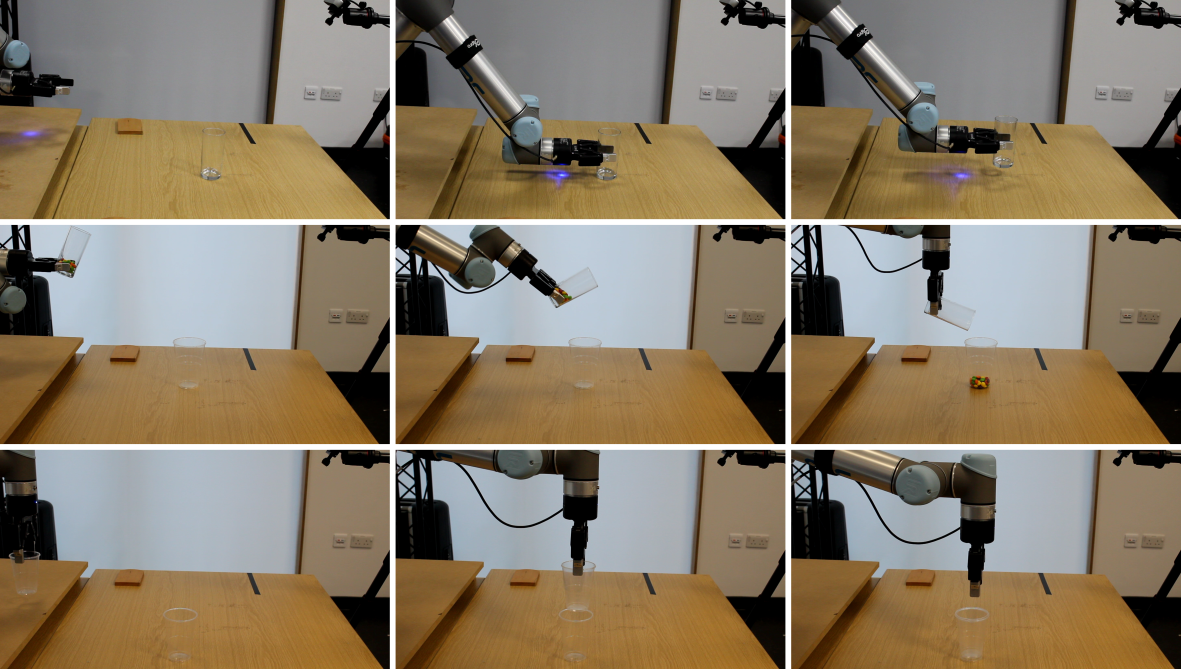}
   \caption{Snapshots of transparent object manipulation. \textbf{First row}: The robotic arm \textbf{picks up} the glass cup. \textbf{Second row}: The robotic arm \textbf{pours} the candies into the disposable cup. \textbf{Third row:k}: The robotic arm \textbf{stacks} one disposable cup into the other cup.}
\label{fig:snapshots}
\end{figure}

The experimental results show that the robot arm can hardly manipulate the transparent objects using the original depth obtained from the RealSense camera. As for the picking task, both our method and ClearGrasp method can significantly improve the success rate. The failure cases are mainly caused by inaccurate depth reconstruction, when no contact edges are detected or surface normals are predicted incorrectly. Nevertheless,  constrained by the depth reconstruction performance, the success rates of the pouring task and the stacking task are not improved impressively with ClearGrasp method.
Compared to ClearGrasp method, our approach with the ability to reconstruct ``contain" regions accurately, can improve the success rates of the stacking task from 57.5\% to 87.5\%, and the pouring task from 15\% to 80\%. 
The snapshots of different manipulation tasks are shown in Fig.~\ref{fig:snapshots}.

\section{Conclusion and Discussion}
\label{Section:6}
In this paper, we introduce a novel framework called A4T that utilizes affordance detection results to achieve multi-step depth reconstruction and manipulation of transparent objects. The extensive experiments show that our multi-step depth reconstruction method is able to reconstruct the depth of regions enclosed by depth discontinued boundaries i.e. the ``contain" regions and shows better performance on our collected \textbf{TRANS-AFF} dataset. The robot manipulation experiments demonstrate that our method is superior to the state-of-the-art method~\cite{sajjan2020clear} in the manipulation tasks related to the affordance ``contain," i.e., the pouring task and the stacking task.

\revised{Our approach is not limited to the three affordance classes used in this work and has potential to be extended to other affordance classes such as a new affordance class ``squeeze" for a dropper. Compared to opaque objects that can be made of a wide variety of materials, transparent objects are limited to materials like glass and plastics. As a consequence, most of them are fragile and tend to have fewer affordance classes. The affordance classes used in our work, i.e., ``contain," ``wrap-grasp" and ``support" represent the majority of the functionalities of the transparent objects in our daily lives and scientific laboratories.}
% \revised{It should be noted that our approach is not limited to the three affordance classes used in this work, but has potential to support other affordance classes. For example, our approach can reconstruct the depth of a transparent knife by using the global optimisation method twice, as the connected edge between the affordance "cut" of the blade and the affordance "wrap-grasp" of the handle is continued. Those three affordance classes are chosen in this work, as they are sufficient for the general manipulation tasks in chemistry laboratories.}

\revised{We validate our approach on structured scenarios that have simple backgrounds and clean tables, as the environments where glasses exist such as chemistry laboratories are normally structured. However, the limitation of scene diversity may constrain the generalisation of our method to unseen challenging environments with complex backgrounds and varying lighting conditions.
We could improve the algorithm generalisation by augmenting the current dataset, e.g., collecting more real-world data with objects under different kinds of backgrounds and lighting conditions, and generating a diverse synthetic affordance dataset with a photo-realistic render engine.}
% such as LuxCoreRender~\cite{LuxCoreRender}.}

\revised{ 
Our approach is an initial step towards reconstructing and manipulating the challenging transparent objects, and is constrained by the prediction of surface normals and contact edges. In cluttered environments, the occlusions between different objects may lead to noisy predictions of surface normals and invisible contact edges, and therefore result in failed depth reconstruction. It could be alleviated by replacing the global optimisation method with an end-to-end reconstruction method that is not dependent on surface normals and contact edges.}

\bibliographystyle{ieeetr}
\bibliography{mybib.bib}

\end{document}